\documentclass[10pt,letterpaper]{article}

\usepackage{cogsci}

\usepackage{tabularx}
\usepackage{graphicx} 
\usepackage{pslatex}
\usepackage{apacite}
\usepackage{algorithm}
\usepackage{algorithmic}
\usepackage{newfloat}
\usepackage{listings}
\usepackage{amssymb}
\usepackage{amsmath}
\usepackage{multirow}
\usepackage{booktabs} 
\usepackage{float} % Roger Levy added this and changed figure/table
\cogscifinalcopy

\title{Improving Brain-to-Image Reconstruction via Fine-Grained Text Bridging}

\author{
{\large \bf Runze Xia, Shuo Feng, Renzhi Wang, Congchi Yin, Xuyun Wen, Piji Li\textbf{$^\ast$}} \\
\texttt{\{xiarunze, fengshuo, rzhwang, congchiyin, wenxuyun, pjli\}}@nuaa.edu.cn\\
 College of Artificial Intelligence, Nanjing University of Aeronautics and Astronautics, Nanjing, China \\
 MIIT Key Laboratory of Pattern Analysis and Machine Intelligence, Nanjing, China \\
 The Key Laboratory of Brain-Machine Intelligence Technology, Ministry of Education, Nanjing, China \\
}

\begin{document}

\maketitle

\begin{abstract}
Brain-to-Image reconstruction aims to recover visual stimuli perceived by humans from brain activity. However, the reconstructed visual stimuli often missing details and semantic inconsistencies, which may be attributed to insufficient semantic information. To address this issue, we propose an approach named Fine-grained Brain-to-Image reconstruction (\textbf{FgB2I}), which employs fine-grained text as bridge to improve image reconstruction. FgB2I comprises three key stages: detail enhancement, decoding fine-grained text descriptions, and text-bridged brain-to-image reconstruction. In the detail-enhancement stage, we leverage large vision–language models to generate fine-grained captions for visual stimuli and experimentally validate its importance. We propose three reward metrics (object accuracy, text-image semantic similarity, and image-image semantic similarity) to guide the language model in decoding fine-grained text descriptions from fMRI signals. The fine-grained text descriptions can be integrated into existing reconstruction methods to achieve fine-grained Brain-to-Image reconstruction.

\end{abstract}
\renewcommand{\thefootnote}{\fnsymbol{footnote}}
\footnotetext[1]{Corresponding author.}
\renewcommand{\thefootnote}{\arabic{footnote}}
\section{Introduction}
Human possesses sophisticated visual and cognitive systems that allow us to easily comprehend visual scenes and guide actions \cite{kandel2000principles, goodale1992separate}. However, the specific mechanisms underlying this remarkable ability still remain unknown, posing a significant challenge in cognitive science, while also captivating the neuroscience community's interest in exploring these mysteries further. The goal of Brain-to-Image reconstruction is to recover visual stimuli perceived by humans from brain activity signals. Recent advances in functional Magnetic Resonance Imaging (fMRI), particularly its high spatial resolution, have driven progress in decoding and reconstructing visual information from brain activity patterns \cite{glover2011overview}.
\begin{figure}[t]
\centering
\includegraphics[scale=0.26]{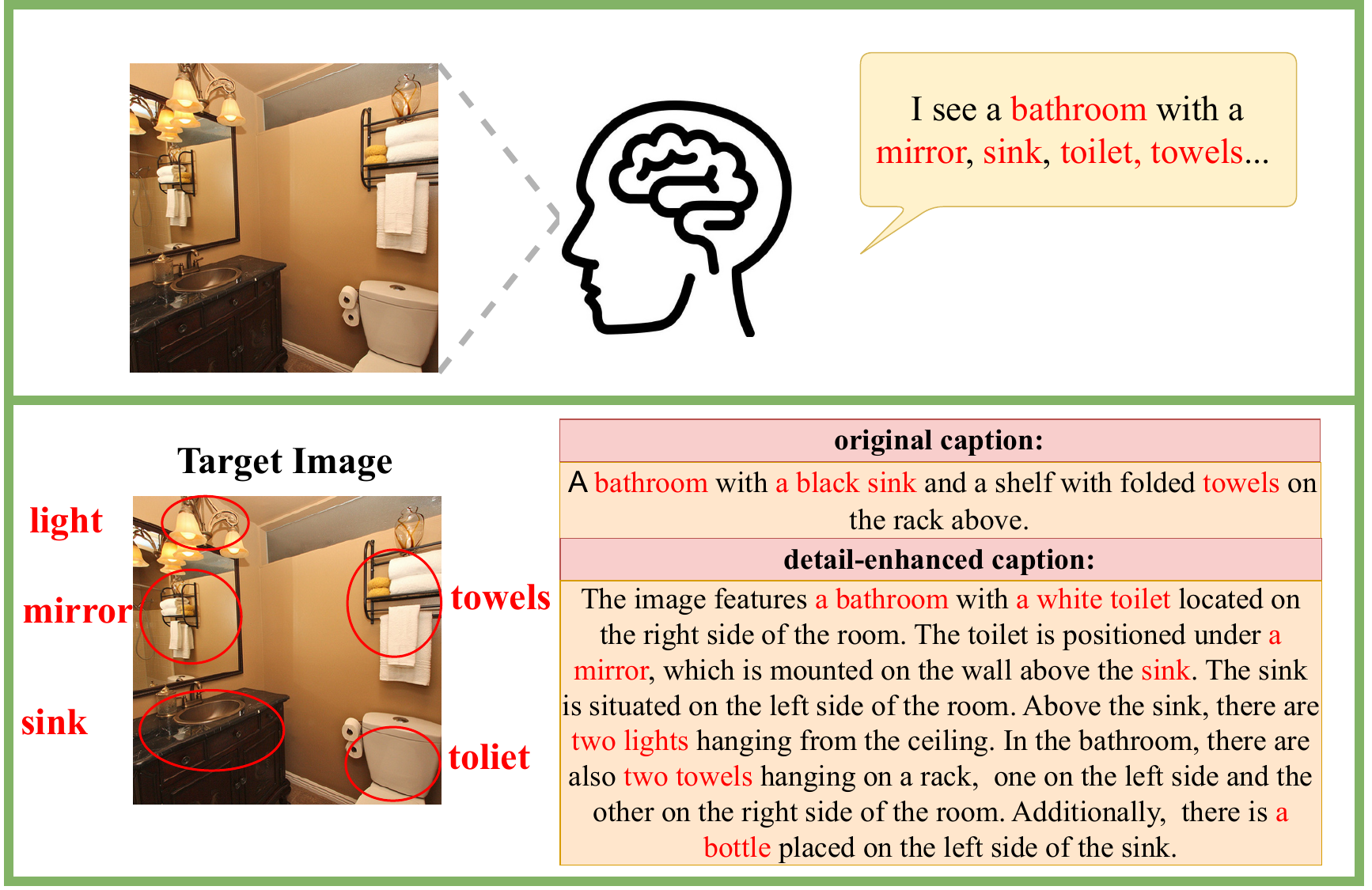}
\caption{The top section is the illustration of cognitive assumptions during scene observation. The bottom section demonstrates a comparison of the granularity between the detail-enhanced and  the original caption. }  
\vspace{-1em}
\label{fig:intro}
\end{figure}

Unlike cameras that record every pixel, the human brain processes visual cognition differently \cite{chen2023rethinking}. It concurrently handles linguistic concepts and semantic information associated with visual scenes \cite{popham2021visual, zhang2020connecting}, as depicted in the top section of Figure \ref{fig:intro}. For example, as shown in the image, when we observe a bathroom scene, our focus may be on the main objects, forming semantic concepts in the brain's representation space such as \textit{mirror}, \textit{sink}, \textit{towels}, and \textit{toilet}. Therefore, semantics plays a vital role in Brain-to-Image reconstruction. Some studies \cite{lu2023minddiffuser, takagi2023high}, have utilized the original image captions from the Natural Scenes Dataset (NSD) \cite{allen2022massive} as semantic targets. These studies use the CLIP \cite{radford2021learning} text encoder to obtain the semantic representation of image captions and train regression models to map fMRI signals to the semantic representation space, thereby achieving semantic reconstruction of images.

However, in more complex visual stimuli, the simple captions often fail to adequately describe the content of the image. As a result, objects observed in the image may not appear in the semantic target, limiting the potential of semantic reconstruction. For instance, the original caption corresponding to the scene in the Figure \ref{fig:intro} is \textit{A bathroom with a black sink and a shelf with folded towels on the rack above}. This caption omits obvious objects like ``toilet'' and ``mirror'', though information about these prominent objects is likely present in the brain activity, leading to a lack of decoding details.

To address this issue, we propose detail-enhancement for image captions using large vision-language models with strong text-image comprehension capabilities to generate fine-grained captions that comprehensively capture all the main information. In the bottom section of Figure \ref{fig:intro}, we illustrate the difference between the two types of captions, showing how the detail-enhanced caption better describes the main content of the image.

Based on this, we propose a novel Brain-to-Image reconstruction method called Fine-grained Brain-to-Image (\textbf{FgB2I}), which aims to improve Brain-to-Image reconstruction by decoding fine-grained semantic descriptions from brain signals as a bridge for image reconstruction. We first train an unified brain-to-text model for all subjects. To decode consistent and fine-grained text descriptions for visual stimuli, we design three reward metrics to guide the model: object accuracy, text-image semantic similarity, and image-image semantic similarity. However, these metrics are non-differentiable and cannot be directly used for model training. Inspired by \cite{DBLP:journals/nature/MnihKSRVBGRFOPB15}, we employ the reinforce algorithm \cite{sutton1999policy} to enable the model to be normally trained using these reward metrics. Through these steps, we are able to decode fine-grained textual descriptions from brain signals. We can easily integrate the decoded fine-grained textual descriptions into the high-level reconstruction of existing methods to improve Brain-to-Image reconstruction.

In summary, our study makes the following contributions:

\begin{itemize}
\item We identify the issue of missing target details in textual descriptions for Brain-to-Image reconstruction and propose the detail-enhancement stage to remedy this issue.
\item  We introduce a novel approach named Fine-grained Brain-to-Image reconstruction (\textbf{FgB2I}), which employs detail-enhanced text as supplements for Brain-to-Image reconstruction. We use three rewards to guide the decoding of text descriptions and employ Reinforce Algorithm to address indifferentiable problems in model co-training stage.
\item We conduct extensive experiments on multiple existing Brain-to-Image reconstruction methods, demonstrating the necessity of detail enhancement and the effectiveness of different reward metrics in our proposed framework. 
\end{itemize}

\begin{figure*}[t]
    \centering
    % \begin{minipage}[t]{\textwidth}
        \includegraphics[scale=0.9]{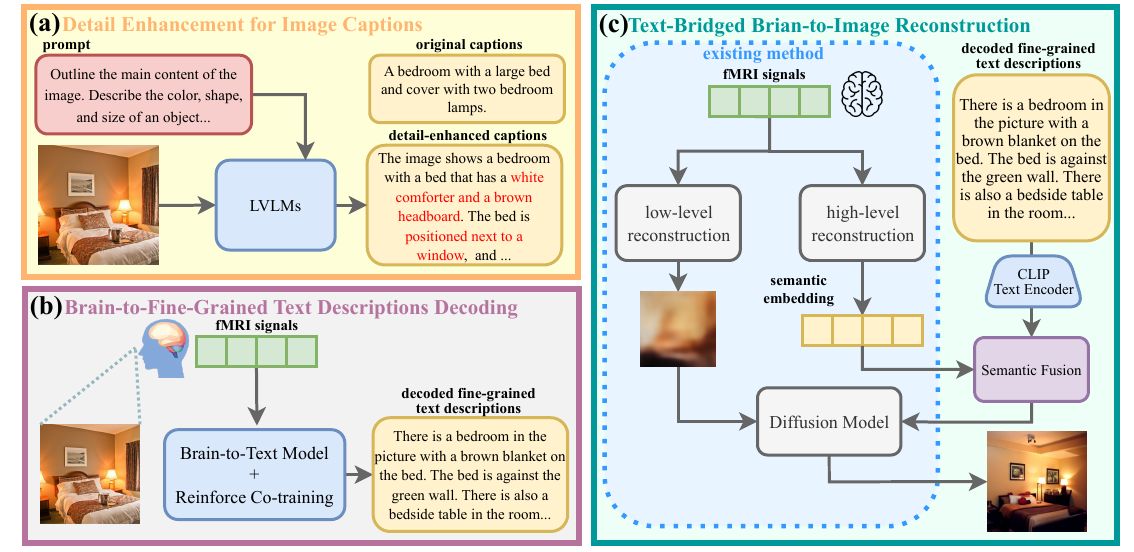}
    % \end{minipage}
    \caption{Overview of FgB2I. \textbf{(a)} Detail enhancement of visual stimuli captions through LVLMs. \textbf{(b)} The process of decoding fine-grained text descriptions from brain signals, with the details inside the blue box further illustrated in Figure \ref{fig:semantic}. \textbf{(c)} The workflow for combining fine-grained text descriptions with existing methods, including semantic fusion through weighted average of text semantic embedding and the integration of text and image embeddings.}  
    \vspace{-1em}
    \label{fig:framework}
\end{figure*}

\section{Related Work}
\subsection{Brain-to-Text Decoding}
Recent advances in deep learning have propelled brain decoding research, particularly in reconstructing language and visual stimuli from cognitive signals. Notable progress includes EEG2Text \cite{wang2022open} achieving open-vocabulary sentence-level decoding from EEG, while UniCoRN \cite{xi2023unicorn} and \textsc{PredFT} \cite{yin2024language} successively advanced fMRI-to-text translation through NLP-inspired architectures and predictive coding mechanisms, highlighting the growing sophistication of Brain-Computer Interfaces.

\subsection{Brain-to-Image Reconstruction}
Visual reconstruction research increasingly adopts diffusion models for their generation capabilities. Key approaches include mapping fMRI data to CLIP text features and Stable Diffusion's VAE latent space \cite{takagi2023high}, and Ozcelik et al.'s two-stage framework \cite{ozcelik2303brain} combining VDVAE \cite{child2021deep} for low-level attributes with Versatile Diffusion \cite{xu2023versatile} guided by CLIP semantics. MindEye \cite{scotti2023reconstructing} further demonstrates how diffusion priors optimize reconstruction accuracy. Selective attention modulates the precision of neural representations during visual encoding by suppressing unattended features' fidelity, while working memory constraints further shape object-level reconstruction through their limited capacity to maintain task-relevant signals during post-perceptual processing stages \cite{chun2011memory,luck1998role,xia2024decoding}.

\section{Method}
\subsection{Detail Enhancement for image captions via LVLMs}
As shown in Figure \ref{fig:framework}(a), we address the limited descriptive capacity of original image captions through visual-language augmentation. Standard captions often omit critical details (e.g., wall paintings and metallic objects in Figure \ref{fig:framework}(a)) that humans naturally perceive. This leads to the loss of information, which the details enhancement process aims to address.

We employ the large visual-language model (LVLM) LLaVA \cite{liu2023visual} to generate fine-grained captions for this stage. LLaVA demonstrates powerful image understanding capabilities, allowing us to extract granular information from images. We employ it with prompt ``\textit{Outline the main content of the image. Describe the color, shape, and size of an object. Use plain language to accurately describe key visual information.}'' to generate text descriptions, effectively capturing object attributes and spatial relationships.

\subsection{Brain-to-Text Decoding Model} \label{sec:b2imodel}
We develop an unified model for all participants to decode text descriptions from their brain signals. Inspired by the concept of prefix tuning \cite{li2021prefix}, we employ a transformer-based network to align fMRI signals with the textual space, obtaining a prefix embedding that guides the GPT2 \cite{radford2019language} model to generate text. Similar to the approach in \cite{scotti2023reconstructing}, we use the predefined template \texttt{nsdgeneral} in the dataset to obtain fMRI signals \( F_i \in \mathbb{R}^{N_s} \)for each participant $s$.

\begin{figure}[t]
\centering
\includegraphics[scale=0.62]{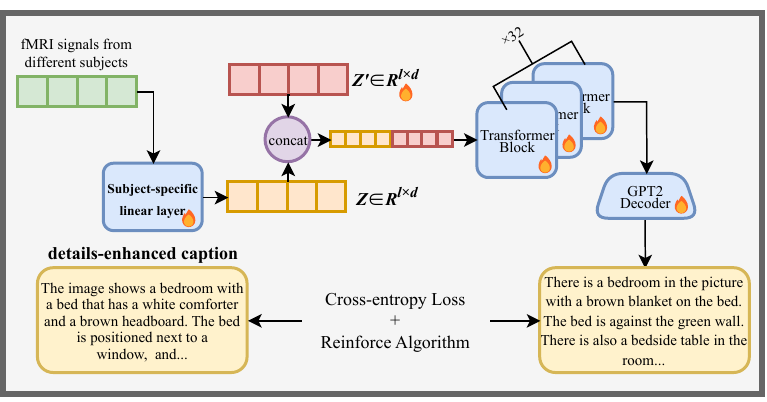}
\caption{Diagram of the brain-to-text model structure and training. The flame represents the trainable components.}  
\vspace{-1em}
\label{fig:semantic}
\end{figure}

The specific process is illustrated in Figure \ref{fig:semantic}. It is important to note that the number of fMRI voxels varies across participants, so we construct a linear layer for each participant to map the fMRI signals to a unified dimensionality. For the fMRI signals of participant $s$, they are first passed through a linear layer to obtain the fMRI embedding \( Z \in \mathbb{R}^{l\times d} \), where $l$ denotes the length of the fMRI prefix and $d$ is the dimension of the word embedding in the GPT2 model. Additionally, a learnable sequence constant embedding \( Z' \in \mathbb{R}^{l\times d} \) is introduced to capture semantic information carried in \( Z \) fully. The \( Z \) and \( Z' \) embedding are concatenated and inputted into a multi-layer Transformer for comprehensive interaction, generating the final fMRI prefix embedding \( p \in \mathbb{R}^{2l \times d}\) as input to the GPT2 model.

\begin{figure*}[t]
    \centering
    % \begin{minipage}[t]{\textwidth}
        \includegraphics[scale=0.65]{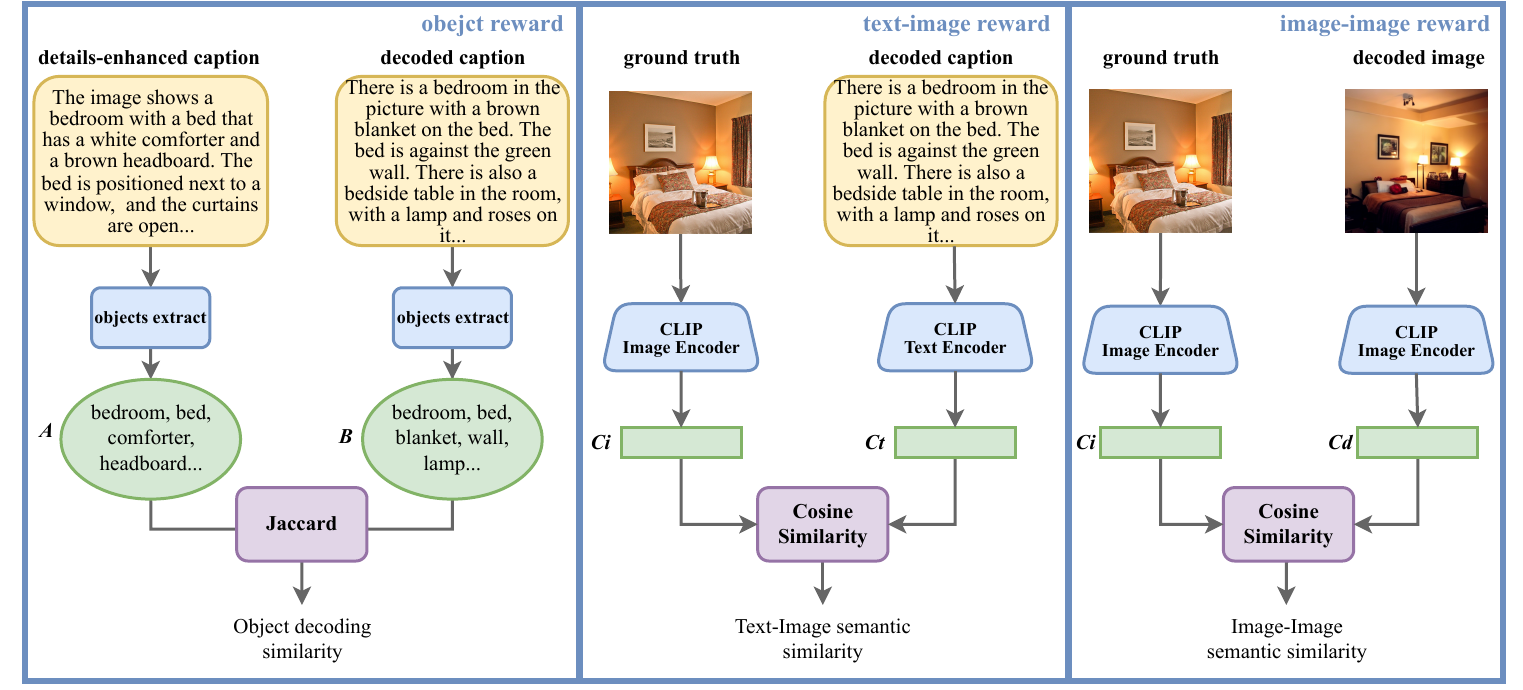}
    % \end{minipage}
    \caption{Three reward function calculation diagrams. \textbf{(Left)} Reward for evaluating the accuracy of decoded objects. \textbf{(Middle)} Semantic similarity between decoded text and visual stimuli, where $C_i$ and $C_t$ represent the corresponding CLIP embedding of the image and text. \textbf{(Right)} Semantic similarity between the reconstructed image and visual stimuli.}  
    \vspace{-1em}
    \label{fig:reward}
\end{figure*}

\begin{figure*}[t]
    \centering
        \includegraphics[scale=0.55]{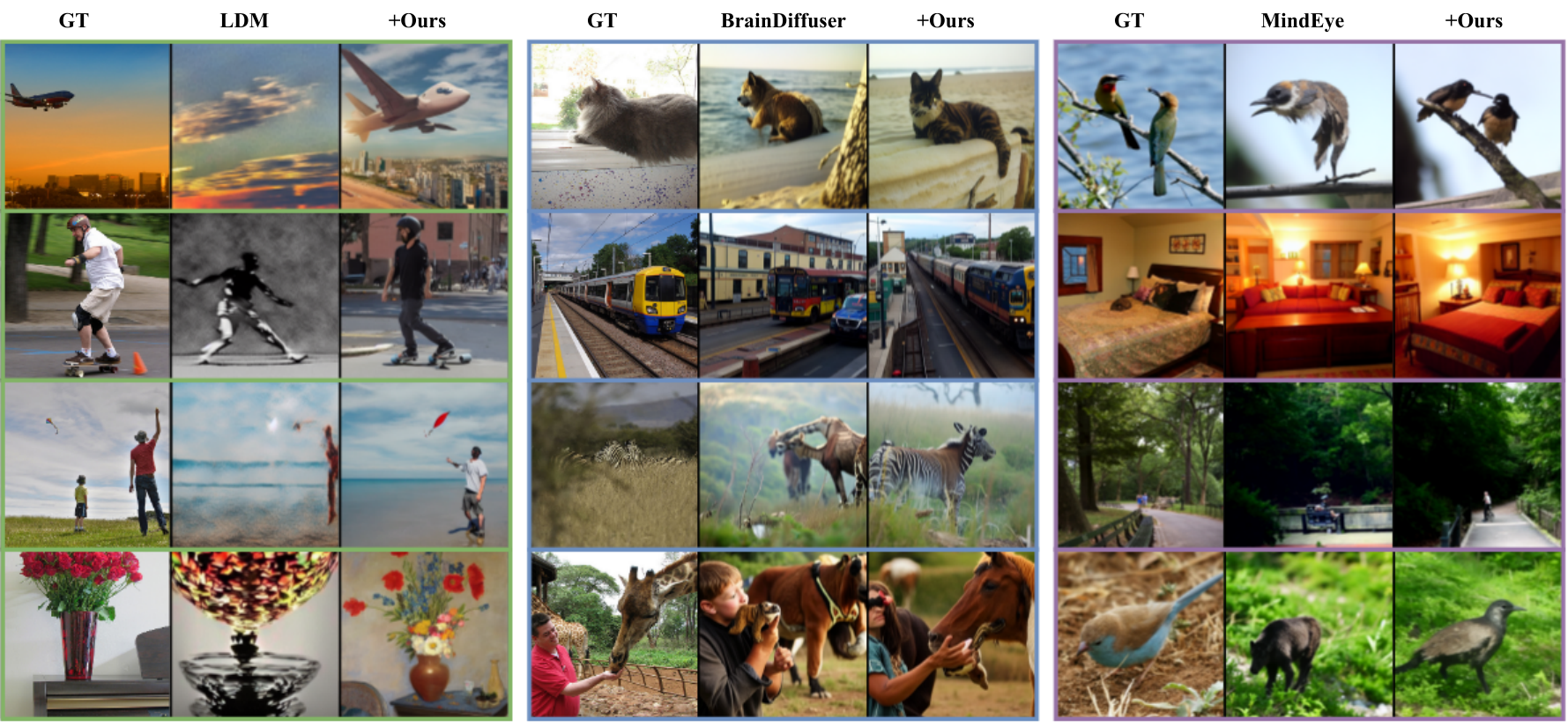}
    \caption{A comparison of the reconstructed image results between existing methods (LDM \cite{takagi2023high}, BrainDiffuser \cite{ozcelik2303brain}, and MindEye \cite{scotti2023reconstructing}) and the results obtained when these methods are combined with our fine-grained text descriptions. GT denotes the corresponding ground truth stimulus image.}  
    \label{fig:results}
\end{figure*}

\subsection{Fine-Grained Text Descriptions Decoding via Reinforced Co-Training} \label{sec:co-training}

To achieve more fine-grained text decoding in the fine-grained semantic reconstruction process, a two-stage co-training approach is employed. In the first stage, we use cross-entropy loss (CE) to train the model to generate text descriptions from brain signals. This stage is crucial for training the GPT2 model to understand and decode the corresponding text descriptions. Here, our goal is to ensure that the decoded text could recognize the objects within the image and to evaluate the semantic similarity between the text and the image. Since the decoded text cannot provide gradient information for model training, we utilize the reinforce algorithm \cite{sutton1999policy} to design a loss function that would achieve these objectives. This algorithm is applied exclusively during the reinforce co-training stage. 

The specific process involves applying softmax to the logits outputted by the GPT2 model to generate a probability distribution $p$. We then sample text sequences based on this distribution, and calculate reward scores for these sequences. These reward scores are used to guide the model towards to generate text that closely match the semantic content of the image. The loss function, which incorporates these reward scores, is formulated as follows:
\begin{equation}
\mathcal{L}(\theta)=\sum_{t=1}^T r_t \log p(a_t|s_t; \theta)
\label{eq:reinforce}
\end{equation}
where $T$ represents the length of the generated text, $a_t$ represents the token sampled at step $t$, $s_t$ represents the corresponding state, $\theta$ represents the model parameters, and $r_t$ represents the reward obtained for the current text. 

Three complementary metrics guide the decoding process (Fig.\ref{fig:reward}): (1) object accuracy, ensuring the inclusion of correct objects; (2) text-image semantic similarity, and (3) image-image semantic similarity.

\paragraph{object accuracy}: The extraction of nouns via TextBlob\footnote{\url{https://textblob.readthedocs.io/en/dev/}} from enhanced/decoded texts, followed by abstract noun filtering (\textit{image}, \textit{photo}, etc.) yields sets A and B. Inter-set congruence is quantified through Jaccard scoring \cite{niwattanakul2013using}:
\begin{equation}
J(A,B) = \frac{|A \cap B|}{|A \cup B|}
\end{equation}
\paragraph{text-image semantic similarity}: The semantic representations of images and textual descriptions $C_i$ and $C_t$ are obtained using CLIP text and image encoders. Subsequently, we measure the similarity between the two representations using cosine similarity:
\begin{equation}
r_{text-image} = \text{Cosine Similarity}(C_i, C_t)
\end{equation}
\paragraph{image-image semantic similarity}: CLIP image-space comparison enforces semantic coherence between reconstructed/original stimuli.

The final loss function is formulated as:
\begin{equation}
\mathcal{L}= \mathcal{L}_{CE} + \alpha \mathcal{L}_1 + \beta \mathcal{L}_2 + \gamma \mathcal{L}_3
\end{equation}
Here, $\alpha$, $\beta$, and $\gamma$ are trade-off factors that balance the importance of each reward in the training process, with $\mathcal{L}_1$, $\mathcal{L}_2$, and $\mathcal{L}_3$ calculated using formula (\ref{eq:reinforce}).
\renewcommand{\arraystretch}{1.0}
\begin{table*}[!t]
    \tiny
    \centering
    \resizebox{\textwidth}{!}{
    \begin{tabular}{lcccc cccc}\toprule
\multirow{2}{*}{Method}&\multicolumn{4}{c}{Low-Level}&\multicolumn{4}{c}{High-Level}\\ \cmidrule(r){2-5}\cmidrule(r){6-9}
&PixCorr$\uparrow$&SSIM$\uparrow$&Alex(2)$\uparrow$&Alex(5)$\uparrow$&Incep$\uparrow$&CLIP$\uparrow$&Eff$\downarrow$&SwAV$\downarrow$\\ \midrule
LDM &/&/&83.0\%&83.0\%&76.0\%&77.0\%&/&/\\ 
LDM+DE&/&/&84.6\%&85.1\%&77.3\%&79.4\%&/&/\\
\midrule
BrainDiffuser&.254&.356&\textbf{94.2\%}&96.2\%&87.2\%&\textbf{91.5\%}&.775&\textbf{.423}\\
BrainDiffuser+DE&\textbf{.255}&\textbf{.357}&93.6\%&\textbf{96.3\%}&\textbf{91.1\%}&\textbf{91.5\%}&\textbf{.737}&\textbf{.423}\\
 \bottomrule 
    \end{tabular}}
    \caption{The reconstruction evaluation of detail-enhancement (denoted as DE in the table) on the LDM \cite{takagi2023high} and BrainDiffuser \cite{ozcelik2303brain} methods, using the same evaluation metrics as theirs. The values presented in the table are the mean of the assessment results from four participants.}
    \vspace{-1em}
    \label{tab:DE}
\end{table*}

\subsection{Text-Bridged Brian-to-Image Reconstruction}

Fine-grained text decoded from brain signals enhances diffusion-model reconstruction via semantic control but can compromise color or structural fidelity. To mitigate this, we integrate our approach into three fMRI-to-image pipelines: LDM, based on the Stable Diffusion model; and BrainDiffuser and MindEye, both built on Versatile Diffusion \cite{takagi2023high, ozcelik2303brain, scotti2023reconstructing, xu2023versatile}.
These frameworks employ two-phase reconstruction:

 \textbf{Low-level.} fMRI signals are mapped to VDVAE/VQVAE latent space \cite{child2020very, van2017neural}, and then decoded into an initial guess image. \textbf{High-level.} Simultaneously, fMRI signals are projected into CLIP space to guide the semantic aspects of image generation.
To balance the semantic information from existing methods with our fine-grained descriptions, we fuse the CLIP embeddings of decoded text with original high-level features. The resulting fused representation replaces the original high-level semantics during image reconstruction.
\renewcommand{\arraystretch}{1.0}
\begin{table*}[!t]
    \tiny
    \centering
    \resizebox{\textwidth}{!}{
    \begin{tabular}{lcccc cccc}\toprule
\multirow{2}{*}{Method}&\multicolumn{4}{c}{Low-Level}&\multicolumn{4}{c}{High-Level}\\ \cmidrule(r){2-5}\cmidrule(r){6-9}
&PixCorr$\uparrow$&SSIM$\uparrow$&Alex(2)$\uparrow$&Alex(5)$\uparrow$&Incep$\uparrow$&CLIP$\uparrow$&Eff$\downarrow$&SwAV$\downarrow$\\ \midrule
LDM &/&/&83.0\%&83.0\%&76.0\%&77.0\%&/&/\\
LDM+Ours&/&/&81.1\%&88.3\%&85.6\%&86.2\%&/&/\\
\midrule
BrainDiffuser&.254&.356&94.2\%&96.2\%&87.2\%&91.5\%&.775&.423\\
BrainDiffuser+Ours&.260&.371&93.9\%&96.4\%&92.4\%&92.2\%&.710&.409\\
\midrule
MindEye&\textbf{.309}&.323&94.7\%&\textbf{97.8\%}&93.8\%&\textbf{94.1\%}&.645&.367\\
MindEye+Ours&.305&\textbf{.354}&\textbf{94.8\%}&\textbf{97.8\%}&\textbf{94.3\%}&93.8\%&\textbf{.637}&\textbf{.360}\\
\bottomrule 
    \end{tabular}}
    \caption{A comparison of the reconstruction results for LDM \cite{takagi2023high}, BrainDiffuser \cite{ozcelik2303brain}, and MindEye \cite{scotti2023reconstructing} methods combined with FgB2I's fine-grained text descriptions, using the same evaluation metrics as theirs. The values presented in the table are the mean of the assessment results from four participants.}
    \vspace{-1em}
    \label{tab:my_label}
\end{table*}
\section{Experimental Settings}
\subsection{Dataset}
In our study, we utilize a subset of the Natural Scenes Dataset (NSD) \cite{allen2022massive}, a comprehensive collection of neuroimaging data obtained from eight participants using a 7-Tesla fMRI scanner.
The dataset encompasses a total of 30–40 scanning sessions, during which each participant viewed three repetitions of 9000-10,000 distinct images sourced from the Microsoft COCO dataset. For our analysis, we focus on data from subjects 1, 2, 5, and 7, who completed the full set of imaging sessions. 982 images are common across all four subjects and are designated as the test dataset, while the remaining trials served as the training dataset. For the test dataset, we average the responses across the three trials associated with each image, whereas for the training dataset, we use the individual trials without averaging.

\subsection{Implementation Details}
In our experiments, we employ the CLIP ViT-B/32 model, and the GPT2-Base model. We utilize an 8-head and 32 layers transformer network, and set the fMRI prefix length $l$ to 10, as well as the maximum length for generated text to 77 (the maximum encoding length of CLIP text encoder). The LLaVA-1.5-7b model is employed for details enhancement for image captions. For quantitative evaluation, we employed a range of metrics, including SSIM \cite{wang2004image}, PixCorr (Pixel-Wise Correlation), Eff \cite{tan1905efficientnet}, SwAV \cite{caron2020unsupervised}, and Two-Way Identification Accuracy (Alexnet \cite{krizhevsky2012imagenet}, and InceptionV3 \cite{szegedy2016rethinking}), following the approach of \cite{scotti2023reconstructing}. SSIM and PixCorr assess low-level similarity, capturing structural and pixel-wise correlations, respectively. Eff and SwAV compute feature distances using pretrained networks to evaluate semantic consistency. Two-Way Identification Accuracy measures whether reconstructed images can be correctly matched to originals based on pretrained networks feature similarities.

During the Brain-to-fine-grained text training phase, we optimize the model using AdamW \cite{loshchilov2019decoupled} with an initial learning rate of 1e-4 over 200 epochs in first stage. The second stage is conducted using the weights of this model from the first stage, with a reduced learning rate of 2e-6 for 10 epochs, maintaining the same other settings, and the parameters $\alpha$, $\beta$, and $\gamma$ are all set to 0.01. 
\begin{table}[!t]
    \small
    % \centering
    % \caption{Caption}
    \begin{tabular}{lcc cc}\toprule
\multirow{2}{*}{Method}&\multicolumn{2}{c}{Low-Level}&\multicolumn{2}{c}{High-Level}\\ \cmidrule(r){2-3}\cmidrule(r){4-5}
&Alex(2)$\uparrow$&Alex(5)$\uparrow$&Incep$\uparrow$&CLIP$\uparrow$\\ \midrule
LDM&77.7\%&75.2\%&67.4\%&70.8\%\\ 
+Ours ($L_{CE}$)&85.1\%&89.3\%&85.4\%&86.1\%\\ 
+Ours ($L_{CE},L_{1}$)&84.8\%&89.6\%&85.9\%&86.9\%\\ 
+Ours ($L_{CE},L_2$)&84.4\%&89.5\%&85.4\%&\textbf{87.1\%}\\ 
+Ours ($L_{CE},L_3$)&85.2\%&89.7\%&\textbf{86.1\%}&86.8\%\\ 
+Ours (all $L$)&\textbf{85.2\%}&\textbf{89.8\%}&85.9\%&86.9\%\\ 
 \bottomrule 
    \end{tabular}
    \caption{The performance of FgB2I on the LDM method using different reward metrics for the loss function.}
    \vspace{-1em}
    \label{tab:ablation}
\end{table}

\section{Results and Analysis}
\subsection{Details Enhancement Results}
To validate the effectiveness of details enhancement, we conduct experiments on two methods: LDM \cite{takagi2023high} and BrainDiffuser \cite{ozcelik2303brain}. In this experiments, we substitute the original image descriptions with detail-enhanced captions generated by LVLMs, while maintaining all other experimental settings and parameters as reported in the original papers.

The results of these experiments are summarized in Table \ref{tab:DE}. The data clearly indicate that incorporating DE leads to improvements across various evaluation metrics, demonstrating that details enhancement is crucial. This suggests that the original captions may lack the necessary granularity to fully capture the intricate details present in the brain's representation of images. However, LVLMs inevitably hallucinate, producing inaccuracies that can sometimes limit the effectiveness of detail enhancement in image captions, though more advanced models can partially mitigate this issue.
\subsection{Main Results}
We apply fine-grained text descriptions to the reconstruction results of a total of three methods, LDM, BrainDiffuser, and MindEye. We first conduct a quantitative analysis and reported the comparison between the original methods and the methods enhanced with our approach, which are presented in Table \ref{tab:my_label}. These results demonstrate the effectiveness of FgB2I in improving brain-to-image reconstruction.

The results clearly show that FgB2I leads to improvements across all methods, with the largest improvement observed in the LDM method, which solely uses text as the semantic control. The other two methods, which use both text and image for semantic control, show more limited improvements. A possible reason is that the combination of text and image semantic conditions in these methods makes the impact of improved text control on image reconstruction more limited.

To visually observe the fine-grained enhancement of image reconstruction by FgB2I, we present a comparison of some reconstructed images with those of the original methods in Figure \ref{fig:results}. It can be visually observed from the figure that incorporating our improvements into various methods leads to more consistent reconstruction outcomes.
\subsection{Ablation Analysis}
During the reinforcement co-training phase for fine-grained text, we train the model using each individual reward loss. To clearly demonstrate the effect of each loss, we conduct the experiments on the LDM method for Subject 1, with the results presented in Table \ref{tab:ablation}. The results show that $L_{1}$ and $L_{2}$ improve semantic alignment, as indicated by higher Incep and CLIP accuracy, but they slightly reduce visual similarity, as reflected in a lower Alex(2) accuracy. $L_{3}$ enhances high-level metrics such as Incep and CLIP while maintaining a good balance with low-level metrics.

\subsection{Case Studies}
The figure \ref{fig:results} illustrates a set of image reconstruction examples, contrasting the existing methods with those refined by incorporating our method for Subject 1. It can be observed from the results that the addition of our method to the LDM approach, which initially exhibited poorer reconstruction quality, yields clearer semantics and improved reconstruction outcomes. When applied to the higher-quality BrainDiffuser and MindEye methods, the integration of our fine-grained text facilitates the correction of semantic inconsistencies, leading to results that more closely align with the ground truth semantics. Furthermore, as evidenced by the third example from the MindEye method, our approach is capable of supplementing fine-grained details, such as including the railings on both sides of the road in our reconstruction, which were omitted in the original method.

\section{Conclusion}
We point out the limitations of using simple captions as semantic reconstruction targets in brain-to-image reconstruction tasks and propose employing LVLM to address this issue. Then, we propose FgB2I, a novel approach that enables the reconstruction of visual stimuli from fMRI signals with enhanced details, and fine-grained text descriptions. Experiments demonstrate FgB2I's capabilities in recovering fine-grained details lost by prior works, showcasing the potential of compensating for insufficient reference details and unifying multi-subject data through semantic decoding. We acknowledge the limitations of current fMRI data and the challenges in decoding neural signals, which inform our future work to improve the granularity of signal decoding. FgB2I provides new perspectives on understanding and reconstructing the intricate process of human visual cognition. 

\section*{Acknowledgements}
This research is supported by the National Natural Science Foundation of China (No.62476127, No.62106105),  the Natural Science Foundation of Jiangsu Province (No.BK20242039), the Basic Research Program of the Bureau of Science and Technology (ILF24001), the Fundamental Research Funds for the Central Universities (No.NJ2023032), the Scientific Research Starting Foundation of Nanjing University of Aeronautics and Astronautics (No.YQR21022), and the High Performance Computing Platform of Nanjing University of Aeronautics and Astronautics.

\bibliographystyle{apacite}

\setlength{\bibleftmargin}{.125in}
\setlength{\bibindent}{-\bibleftmargin}

\bibliography{CogSci_Template}

\begin{thebibliography}{}

\bibitem [\protect \citeauthoryear {%
Allen%
\ \protect \BOthers {.}}{%
Allen%
\ \protect \BOthers {.}}{%
{\protect \APACyear {2022}}%
}]{%
allen2022massive}
\APACinsertmetastar {%
allen2022massive}%
\begin{APACrefauthors}%
Allen, E\BPBI J.%
, St-Yves, G.%
, Wu, Y.%
, Breedlove, J\BPBI L.%
, Prince, J\BPBI S.%
, Dowdle, L\BPBI T.%
\BDBL {}others%
\end{APACrefauthors}%
\unskip\
\newblock
\APACrefYearMonthDay{2022}{}{}.
\newblock
{\BBOQ}\APACrefatitle {A massive 7T fMRI dataset to bridge cognitive neuroscience and artificial intelligence} {A massive 7t fmri dataset to bridge cognitive neuroscience and artificial intelligence}.{\BBCQ}
\newblock
\APACjournalVolNumPages{Nature neuroscience}{25}{1}{116--126}.
\PrintBackRefs{\CurrentBib}

\bibitem [\protect \citeauthoryear {%
Caron%
\ \protect \BOthers {.}}{%
Caron%
\ \protect \BOthers {.}}{%
{\protect \APACyear {2020}}%
}]{%
caron2020unsupervised}
\APACinsertmetastar {%
caron2020unsupervised}%
\begin{APACrefauthors}%
Caron, M.%
, Misra, I.%
, Mairal, J.%
, Goyal, P.%
, Bojanowski, P.%
\BCBL {}\ \BBA {} Joulin, A.%
\end{APACrefauthors}%
\unskip\
\newblock
\APACrefYearMonthDay{2020}{}{}.
\newblock
{\BBOQ}\APACrefatitle {Unsupervised learning of visual features by contrasting cluster assignments} {Unsupervised learning of visual features by contrasting cluster assignments}.{\BBCQ}
\newblock
\APACjournalVolNumPages{Advances in neural information processing systems}{33}{}{9912--9924}.
\PrintBackRefs{\CurrentBib}

\bibitem [\protect \citeauthoryear {%
Chen%
, Qi%
\BCBL {}\ \BBA {} Pan%
}{%
Chen%
\ \protect \BOthers {.}}{%
{\protect \APACyear {2023}}%
}]{%
chen2023rethinking}
\APACinsertmetastar {%
chen2023rethinking}%
\begin{APACrefauthors}%
Chen, J.%
, Qi, Y.%
\BCBL {}\ \BBA {} Pan, G.%
\end{APACrefauthors}%
\unskip\
\newblock
\APACrefYearMonthDay{2023}{}{}.
\newblock
{\BBOQ}\APACrefatitle {Rethinking Visual Reconstruction: Experience-Based Content Completion Guided by Visual Cues} {Rethinking visual reconstruction: Experience-based content completion guided by visual cues}.{\BBCQ}
\newblock
\APACjournalVolNumPages{}{202}{}{4856--4866}.
\newblock
\begin{APACrefURL} \url{https://proceedings.mlr.press/v202/chen23v.html} \end{APACrefURL}
\PrintBackRefs{\CurrentBib}

\bibitem [\protect \citeauthoryear {%
Child%
}{%
Child%
}{%
{\protect \APACyear {2020}}%
}]{%
child2020very}
\APACinsertmetastar {%
child2020very}%
\begin{APACrefauthors}%
Child, R.%
\end{APACrefauthors}%
\unskip\
\newblock
\APACrefYearMonthDay{2020}{}{}.
\newblock
{\BBOQ}\APACrefatitle {Very deep vaes generalize autoregressive models and can outperform them on images} {Very deep vaes generalize autoregressive models and can outperform them on images}.{\BBCQ}
\newblock
\APACjournalVolNumPages{arXiv preprint arXiv:2011.10650}{}{}{}.
\PrintBackRefs{\CurrentBib}

\bibitem [\protect \citeauthoryear {%
Child%
}{%
Child%
}{%
{\protect \APACyear {2021}}%
}]{%
child2021deep}
\APACinsertmetastar {%
child2021deep}%
\begin{APACrefauthors}%
Child, R.%
\end{APACrefauthors}%
\unskip\
\newblock
\APACrefYearMonthDay{2021}{}{}.
\newblock
{\BBOQ}\APACrefatitle {Very Deep VAEs Generalize Autoregressive Models and Can Outperform Them on Images} {Very deep vaes generalize autoregressive models and can outperform them on images}.{\BBCQ}
\newblock
\BIn{} \APACrefbtitle {9th International Conference on Learning Representations, {ICLR} 2021, Virtual Event, Austria, May 3-7, 2021.} {9th international conference on learning representations, {ICLR} 2021, virtual event, austria, may 3-7, 2021.}
\newblock
\APACaddressPublisher{}{OpenReview.net}.
\newblock
\begin{APACrefURL} \url{https://openreview.net/forum?id=RLRXCV6DbEJ} \end{APACrefURL}
\PrintBackRefs{\CurrentBib}

\bibitem [\protect \citeauthoryear {%
Chun%
\ \BBA {} Johnson%
}{%
Chun%
\ \BBA {} Johnson%
}{%
{\protect \APACyear {2011}}%
}]{%
chun2011memory}
\APACinsertmetastar {%
chun2011memory}%
\begin{APACrefauthors}%
Chun, M\BPBI M.%
\BCBT {}\ \BBA {} Johnson, M\BPBI K.%
\end{APACrefauthors}%
\unskip\
\newblock
\APACrefYearMonthDay{2011}{}{}.
\newblock
{\BBOQ}\APACrefatitle {Memory: Enduring traces of perceptual and reflective attention} {Memory: Enduring traces of perceptual and reflective attention}.{\BBCQ}
\newblock
\APACjournalVolNumPages{Neuron}{72}{4}{520--535}.
\PrintBackRefs{\CurrentBib}

\bibitem [\protect \citeauthoryear {%
Glover%
}{%
Glover%
}{%
{\protect \APACyear {2011}}%
}]{%
glover2011overview}
\APACinsertmetastar {%
glover2011overview}%
\begin{APACrefauthors}%
Glover, G\BPBI H.%
\end{APACrefauthors}%
\unskip\
\newblock
\APACrefYearMonthDay{2011}{}{}.
\newblock
{\BBOQ}\APACrefatitle {Overview of functional magnetic resonance imaging} {Overview of functional magnetic resonance imaging}.{\BBCQ}
\newblock
\APACjournalVolNumPages{Neurosurgery Clinics}{22}{2}{133--139}.
\PrintBackRefs{\CurrentBib}

\bibitem [\protect \citeauthoryear {%
Goodale%
\ \BBA {} Milner%
}{%
Goodale%
\ \BBA {} Milner%
}{%
{\protect \APACyear {1992}}%
}]{%
goodale1992separate}
\APACinsertmetastar {%
goodale1992separate}%
\begin{APACrefauthors}%
Goodale, M\BPBI A.%
\BCBT {}\ \BBA {} Milner, A\BPBI D.%
\end{APACrefauthors}%
\unskip\
\newblock
\APACrefYearMonthDay{1992}{}{}.
\newblock
{\BBOQ}\APACrefatitle {Separate visual pathways for perception and action} {Separate visual pathways for perception and action}.{\BBCQ}
\newblock
\APACjournalVolNumPages{Trends in neurosciences}{15}{1}{20--25}.
\PrintBackRefs{\CurrentBib}

\bibitem [\protect \citeauthoryear {%
Kandel%
\ \protect \BOthers {.}}{%
Kandel%
\ \protect \BOthers {.}}{%
{\protect \APACyear {2000}}%
}]{%
kandel2000principles}
\APACinsertmetastar {%
kandel2000principles}%
\begin{APACrefauthors}%
Kandel, E\BPBI R.%
, Schwartz, J\BPBI H.%
, Jessell, T\BPBI M.%
, Siegelbaum, S.%
, Hudspeth, A\BPBI J.%
, Mack, S.%
\BCBL {}\ \BOthersPeriod {.}\end{APACrefauthors}%
\unskip\
\newblock
\APACrefYear{2000}.
\newblock
\APACrefbtitle {Principles of neural science} {Principles of neural science}\ (\BVOL~4).
\newblock
\APACaddressPublisher{}{McGraw-hill New York}.
\PrintBackRefs{\CurrentBib}

\bibitem [\protect \citeauthoryear {%
Krizhevsky%
, Sutskever%
\BCBL {}\ \BBA {} Hinton%
}{%
Krizhevsky%
\ \protect \BOthers {.}}{%
{\protect \APACyear {2012}}%
}]{%
krizhevsky2012imagenet}
\APACinsertmetastar {%
krizhevsky2012imagenet}%
\begin{APACrefauthors}%
Krizhevsky, A.%
, Sutskever, I.%
\BCBL {}\ \BBA {} Hinton, G\BPBI E.%
\end{APACrefauthors}%
\unskip\
\newblock
\APACrefYearMonthDay{2012}{}{}.
\newblock
{\BBOQ}\APACrefatitle {Imagenet classification with deep convolutional neural networks} {Imagenet classification with deep convolutional neural networks}.{\BBCQ}
\newblock
\BIn{} (\BVOL~25).
\PrintBackRefs{\CurrentBib}

\bibitem [\protect \citeauthoryear {%
Li%
\ \BBA {} Liang%
}{%
Li%
\ \BBA {} Liang%
}{%
{\protect \APACyear {2021}}%
}]{%
li2021prefix}
\APACinsertmetastar {%
li2021prefix}%
\begin{APACrefauthors}%
Li, X\BPBI L.%
\BCBT {}\ \BBA {} Liang, P.%
\end{APACrefauthors}%
\unskip\
\newblock
\APACrefYearMonthDay{2021}{}{}.
\newblock
{\BBOQ}\APACrefatitle {Prefix-tuning: Optimizing continuous prompts for generation} {Prefix-tuning: Optimizing continuous prompts for generation}.{\BBCQ}
\newblock
\APACjournalVolNumPages{arXiv preprint arXiv:2101.00190}{}{}{}.
\PrintBackRefs{\CurrentBib}

\bibitem [\protect \citeauthoryear {%
Liu%
, Li%
, Wu%
\BCBL {}\ \BBA {} Lee%
}{%
Liu%
\ \protect \BOthers {.}}{%
{\protect \APACyear {2023}}%
}]{%
liu2023visual}
\APACinsertmetastar {%
liu2023visual}%
\begin{APACrefauthors}%
Liu, H.%
, Li, C.%
, Wu, Q.%
\BCBL {}\ \BBA {} Lee, Y\BPBI J.%
\end{APACrefauthors}%
\unskip\
\newblock
\APACrefYearMonthDay{2023}{}{}.
\newblock
{\BBOQ}\APACrefatitle {Visual Instruction Tuning} {Visual instruction tuning}.{\BBCQ}
\newblock
\APACjournalVolNumPages{CoRR}{abs/2304.08485}{}{}.
\newblock
\begin{APACrefURL} \url{https://doi.org/10.48550/arXiv.2304.08485} \end{APACrefURL}
\newblock
\begin{APACrefDOI} \doi{10.48550/ARXIV.2304.08485} \end{APACrefDOI}
\PrintBackRefs{\CurrentBib}

\bibitem [\protect \citeauthoryear {%
Loshchilov%
\ \BBA {} Hutter%
}{%
Loshchilov%
\ \BBA {} Hutter%
}{%
{\protect \APACyear {2019}}%
}]{%
loshchilov2019decoupled}
\APACinsertmetastar {%
loshchilov2019decoupled}%
\begin{APACrefauthors}%
Loshchilov, I.%
\BCBT {}\ \BBA {} Hutter, F.%
\end{APACrefauthors}%
\unskip\
\newblock
\APACrefYearMonthDay{2019}{}{}.
\newblock
{\BBOQ}\APACrefatitle {Decoupled Weight Decay Regularization} {Decoupled weight decay regularization}.{\BBCQ}
\newblock
\BIn{} \APACrefbtitle {7th International Conference on Learning Representations, {ICLR} 2019, New Orleans, LA, USA, May 6-9, 2019.} {7th international conference on learning representations, {ICLR} 2019, new orleans, la, usa, may 6-9, 2019.}
\newblock
\APACaddressPublisher{}{OpenReview.net}.
\newblock
\begin{APACrefURL} \url{https://openreview.net/forum?id=Bkg6RiCqY7} \end{APACrefURL}
\PrintBackRefs{\CurrentBib}

\bibitem [\protect \citeauthoryear {%
Lu%
, Du%
, Zhou%
, Wang%
\BCBL {}\ \BBA {} He%
}{%
Lu%
\ \protect \BOthers {.}}{%
{\protect \APACyear {2023}}%
}]{%
lu2023minddiffuser}
\APACinsertmetastar {%
lu2023minddiffuser}%
\begin{APACrefauthors}%
Lu, Y.%
, Du, C.%
, Zhou, Q.%
, Wang, D.%
\BCBL {}\ \BBA {} He, H.%
\end{APACrefauthors}%
\unskip\
\newblock
\APACrefYearMonthDay{2023}{}{}.
\newblock
{\BBOQ}\APACrefatitle {MindDiffuser: Controlled Image Reconstruction from Human Brain Activity with Semantic and Structural Diffusion} {Minddiffuser: Controlled image reconstruction from human brain activity with semantic and structural diffusion}.{\BBCQ}
\newblock
\BIn{} A.~El{-}Saddik\ \BOthers {.}\ (\BEDS), \APACrefbtitle {Proceedings of the 31st {ACM} International Conference on Multimedia, {MM} 2023, Ottawa, ON, Canada, 29 October 2023- 3 November 2023} {Proceedings of the 31st {ACM} international conference on multimedia, {MM} 2023, ottawa, on, canada, 29 october 2023- 3 november 2023}\ (\BPGS\ 5899--5908).
\newblock
\APACaddressPublisher{}{{ACM}}.
\newblock
\begin{APACrefURL} \url{https://doi.org/10.1145/3581783.3613832} \end{APACrefURL}
\newblock
\begin{APACrefDOI} \doi{10.1145/3581783.3613832} \end{APACrefDOI}
\PrintBackRefs{\CurrentBib}

\bibitem [\protect \citeauthoryear {%
Luck%
\ \BBA {} Ford%
}{%
Luck%
\ \BBA {} Ford%
}{%
{\protect \APACyear {1998}}%
}]{%
luck1998role}
\APACinsertmetastar {%
luck1998role}%
\begin{APACrefauthors}%
Luck, S\BPBI J.%
\BCBT {}\ \BBA {} Ford, M\BPBI A.%
\end{APACrefauthors}%
\unskip\
\newblock
\APACrefYearMonthDay{1998}{}{}.
\newblock
{\BBOQ}\APACrefatitle {On the role of selective attention in visual perception} {On the role of selective attention in visual perception}.{\BBCQ}
\newblock
\APACjournalVolNumPages{Proceedings of the National Academy of Sciences}{95}{3}{825--830}.
\PrintBackRefs{\CurrentBib}

\bibitem [\protect \citeauthoryear {%
Mnih%
\ \protect \BOthers {.}}{%
Mnih%
\ \protect \BOthers {.}}{%
{\protect \APACyear {2015}}%
}]{%
DBLP:journals/nature/MnihKSRVBGRFOPB15}
\APACinsertmetastar {%
DBLP:journals/nature/MnihKSRVBGRFOPB15}%
\begin{APACrefauthors}%
Mnih, V.%
, Kavukcuoglu, K.%
, Silver, D.%
, Rusu, A\BPBI A.%
, Veness, J.%
, Bellemare, M\BPBI G.%
\BDBL {}Hassabis, D.%
\end{APACrefauthors}%
\unskip\
\newblock
\APACrefYearMonthDay{2015}{}{}.
\newblock
{\BBOQ}\APACrefatitle {Human-level control through deep reinforcement learning} {Human-level control through deep reinforcement learning}.{\BBCQ}
\newblock
\APACjournalVolNumPages{Nat.}{518}{7540}{529--533}.
\newblock
\begin{APACrefURL} \url{https://doi.org/10.1038/nature14236} \end{APACrefURL}
\newblock
\begin{APACrefDOI} \doi{10.1038/NATURE14236} \end{APACrefDOI}
\PrintBackRefs{\CurrentBib}

\bibitem [\protect \citeauthoryear {%
Niwattanakul%
, Singthongchai%
, Naenudorn%
\BCBL {}\ \BBA {} Wanapu%
}{%
Niwattanakul%
\ \protect \BOthers {.}}{%
{\protect \APACyear {2013}}%
}]{%
niwattanakul2013using}
\APACinsertmetastar {%
niwattanakul2013using}%
\begin{APACrefauthors}%
Niwattanakul, S.%
, Singthongchai, J.%
, Naenudorn, E.%
\BCBL {}\ \BBA {} Wanapu, S.%
\end{APACrefauthors}%
\unskip\
\newblock
\APACrefYearMonthDay{2013}{}{}.
\newblock
{\BBOQ}\APACrefatitle {Using of Jaccard coefficient for keywords similarity} {Using of jaccard coefficient for keywords similarity}.{\BBCQ}
\newblock
\BIn{} \APACrefbtitle {Proceedings of the international multiconference of engineers and computer scientists} {Proceedings of the international multiconference of engineers and computer scientists}\ (\BVOL~1, \BPGS\ 380--384).
\PrintBackRefs{\CurrentBib}

\bibitem [\protect \citeauthoryear {%
Ozcelik%
\ \BBA {} VanRullen%
}{%
Ozcelik%
\ \BBA {} VanRullen%
}{%
{\protect \APACyear {2023}}%
}]{%
ozcelik2303brain}
\APACinsertmetastar {%
ozcelik2303brain}%
\begin{APACrefauthors}%
Ozcelik, F.%
\BCBT {}\ \BBA {} VanRullen, R.%
\end{APACrefauthors}%
\unskip\
\newblock
\APACrefYearMonthDay{2023}{}{}.
\newblock
{\BBOQ}\APACrefatitle {Natural scene reconstruction from fMRI signals using generative latent diffusion} {Natural scene reconstruction from fmri signals using generative latent diffusion}.{\BBCQ}
\newblock
\APACjournalVolNumPages{Scientific Reports}{13}{1}{15666}.
\PrintBackRefs{\CurrentBib}

\bibitem [\protect \citeauthoryear {%
Popham%
\ \protect \BOthers {.}}{%
Popham%
\ \protect \BOthers {.}}{%
{\protect \APACyear {2021}}%
}]{%
popham2021visual}
\APACinsertmetastar {%
popham2021visual}%
\begin{APACrefauthors}%
Popham, S\BPBI F.%
, Huth, A\BPBI G.%
, Bilenko, N\BPBI Y.%
, Deniz, F.%
, Gao, J\BPBI S.%
, Nunez-Elizalde, A\BPBI O.%
\BCBL {}\ \BBA {} Gallant, J\BPBI L.%
\end{APACrefauthors}%
\unskip\
\newblock
\APACrefYearMonthDay{2021}{}{}.
\newblock
{\BBOQ}\APACrefatitle {Visual and linguistic semantic representations are aligned at the border of human visual cortex} {Visual and linguistic semantic representations are aligned at the border of human visual cortex}.{\BBCQ}
\newblock
\APACjournalVolNumPages{Nature neuroscience}{24}{11}{1628--1636}.
\PrintBackRefs{\CurrentBib}

\bibitem [\protect \citeauthoryear {%
Radford%
\ \protect \BOthers {.}}{%
Radford%
\ \protect \BOthers {.}}{%
{\protect \APACyear {2021}}%
}]{%
radford2021learning}
\APACinsertmetastar {%
radford2021learning}%
\begin{APACrefauthors}%
Radford, A.%
, Kim, J\BPBI W.%
, Hallacy, C.%
, Ramesh, A.%
, Goh, G.%
, Agarwal, S.%
\BDBL {}others%
\end{APACrefauthors}%
\unskip\
\newblock
\APACrefYearMonthDay{2021}{}{}.
\newblock
{\BBOQ}\APACrefatitle {Learning transferable visual models from natural language supervision} {Learning transferable visual models from natural language supervision}.{\BBCQ}
\newblock
\BIn{} \APACrefbtitle {International conference on machine learning} {International conference on machine learning}\ (\BPGS\ 8748--8763).
\PrintBackRefs{\CurrentBib}

\bibitem [\protect \citeauthoryear {%
Radford%
\ \protect \BOthers {.}}{%
Radford%
\ \protect \BOthers {.}}{%
{\protect \APACyear {2019}}%
}]{%
radford2019language}
\APACinsertmetastar {%
radford2019language}%
\begin{APACrefauthors}%
Radford, A.%
, Wu, J.%
, Child, R.%
, Luan, D.%
, Amodei, D.%
, Sutskever, I.%
\BCBL {}\ \BOthersPeriod {.}\end{APACrefauthors}%
\unskip\
\newblock
\APACrefYearMonthDay{2019}{}{}.
\newblock
{\BBOQ}\APACrefatitle {Language models are unsupervised multitask learners} {Language models are unsupervised multitask learners}.{\BBCQ}
\newblock
\APACjournalVolNumPages{OpenAI blog}{1}{8}{9}.
\PrintBackRefs{\CurrentBib}

\bibitem [\protect \citeauthoryear {%
Scotti%
\ \protect \BOthers {.}}{%
Scotti%
\ \protect \BOthers {.}}{%
{\protect \APACyear {2023}}%
}]{%
scotti2023reconstructing}
\APACinsertmetastar {%
scotti2023reconstructing}%
\begin{APACrefauthors}%
Scotti, P\BPBI S.%
, Banerjee, A.%
, Goode, J.%
, Shabalin, S.%
, Nguyen, A.%
, Cohen, E.%
\BDBL {}Abraham, T\BPBI M.%
\end{APACrefauthors}%
\unskip\
\newblock
\APACrefYearMonthDay{2023}{}{}.
\newblock
{\BBOQ}\APACrefatitle {Reconstructing the Mind's Eye: fMRI-to-Image with Contrastive Learning and Diffusion Priors} {Reconstructing the mind's eye: fmri-to-image with contrastive learning and diffusion priors}.{\BBCQ}
\newblock
\APACjournalVolNumPages{CoRR}{abs/2305.18274}{}{}.
\newblock
\begin{APACrefURL} \url{https://doi.org/10.48550/arXiv.2305.18274} \end{APACrefURL}
\newblock
\begin{APACrefDOI} \doi{10.48550/ARXIV.2305.18274} \end{APACrefDOI}
\PrintBackRefs{\CurrentBib}

\bibitem [\protect \citeauthoryear {%
Sutton%
, McAllester%
, Singh%
\BCBL {}\ \BBA {} Mansour%
}{%
Sutton%
\ \protect \BOthers {.}}{%
{\protect \APACyear {1999}}%
}]{%
sutton1999policy}
\APACinsertmetastar {%
sutton1999policy}%
\begin{APACrefauthors}%
Sutton, R\BPBI S.%
, McAllester, D.%
, Singh, S.%
\BCBL {}\ \BBA {} Mansour, Y.%
\end{APACrefauthors}%
\unskip\
\newblock
\APACrefYearMonthDay{1999}{}{}.
\newblock
{\BBOQ}\APACrefatitle {Policy gradient methods for reinforcement learning with function approximation} {Policy gradient methods for reinforcement learning with function approximation}.{\BBCQ}
\newblock
\APACjournalVolNumPages{Advances in neural information processing systems}{12}{}{}.
\PrintBackRefs{\CurrentBib}

\bibitem [\protect \citeauthoryear {%
Szegedy%
, Vanhoucke%
, Ioffe%
, Shlens%
\BCBL {}\ \BBA {} Wojna%
}{%
Szegedy%
\ \protect \BOthers {.}}{%
{\protect \APACyear {2016}}%
}]{%
szegedy2016rethinking}
\APACinsertmetastar {%
szegedy2016rethinking}%
\begin{APACrefauthors}%
Szegedy, C.%
, Vanhoucke, V.%
, Ioffe, S.%
, Shlens, J.%
\BCBL {}\ \BBA {} Wojna, Z.%
\end{APACrefauthors}%
\unskip\
\newblock
\APACrefYearMonthDay{2016}{}{}.
\newblock
{\BBOQ}\APACrefatitle {Rethinking the Inception Architecture for Computer Vision} {Rethinking the inception architecture for computer vision}.{\BBCQ}
\newblock
\BIn{} \APACrefbtitle {2016 {IEEE} Conference on Computer Vision and Pattern Recognition, {CVPR} 2016, Las Vegas, NV, USA, June 27-30, 2016} {2016 {IEEE} conference on computer vision and pattern recognition, {CVPR} 2016, las vegas, nv, usa, june 27-30, 2016}\ (\BPGS\ 2818--2826).
\newblock
\APACaddressPublisher{}{{IEEE} Computer Society}.
\newblock
\begin{APACrefURL} \url{https://doi.org/10.1109/CVPR.2016.308} \end{APACrefURL}
\newblock
\begin{APACrefDOI} \doi{10.1109/CVPR.2016.308} \end{APACrefDOI}
\PrintBackRefs{\CurrentBib}

\bibitem [\protect \citeauthoryear {%
Takagi%
\ \BBA {} Nishimoto%
}{%
Takagi%
\ \BBA {} Nishimoto%
}{%
{\protect \APACyear {2023}}%
}]{%
takagi2023high}
\APACinsertmetastar {%
takagi2023high}%
\begin{APACrefauthors}%
Takagi, Y.%
\BCBT {}\ \BBA {} Nishimoto, S.%
\end{APACrefauthors}%
\unskip\
\newblock
\APACrefYearMonthDay{2023}{}{}.
\newblock
{\BBOQ}\APACrefatitle {High-resolution image reconstruction with latent diffusion models from human brain activity} {High-resolution image reconstruction with latent diffusion models from human brain activity}.{\BBCQ}
\newblock
\BIn{} \APACrefbtitle {{IEEE/CVF} Conference on Computer Vision and Pattern Recognition, {CVPR} 2023, Vancouver, BC, Canada, June 17-24, 2023} {{IEEE/CVF} conference on computer vision and pattern recognition, {CVPR} 2023, vancouver, bc, canada, june 17-24, 2023}\ (\BPGS\ 14453--14463).
\newblock
\APACaddressPublisher{}{{IEEE}}.
\newblock
\begin{APACrefURL} \url{https://doi.org/10.1109/CVPR52729.2023.01389} \end{APACrefURL}
\newblock
\begin{APACrefDOI} \doi{10.1109/CVPR52729.2023.01389} \end{APACrefDOI}
\PrintBackRefs{\CurrentBib}

\bibitem [\protect \citeauthoryear {%
Tan%
\ \BBA {} Le%
}{%
Tan%
\ \BBA {} Le%
}{%
{\protect \APACyear {2019}}%
}]{%
tan1905efficientnet}
\APACinsertmetastar {%
tan1905efficientnet}%
\begin{APACrefauthors}%
Tan, M.%
\BCBT {}\ \BBA {} Le, Q\BPBI V.%
\end{APACrefauthors}%
\unskip\
\newblock
\APACrefYearMonthDay{2019}{}{}.
\newblock
{\BBOQ}\APACrefatitle {EfficientNet: Rethinking Model Scaling for Convolutional Neural Networks} {Efficientnet: Rethinking model scaling for convolutional neural networks}.{\BBCQ}
\newblock
\BIn{} K.~Chaudhuri\ \BBA {} R.~Salakhutdinov\ (\BEDS), \APACrefbtitle {Proceedings of the 36th International Conference on Machine Learning, {ICML} 2019, 9-15 June 2019, Long Beach, California, {USA}} {Proceedings of the 36th international conference on machine learning, {ICML} 2019, 9-15 june 2019, long beach, california, {USA}}\ (\BVOL~97, \BPGS\ 6105--6114).
\newblock
\APACaddressPublisher{}{{PMLR}}.
\newblock
\begin{APACrefURL} \url{http://proceedings.mlr.press/v97/tan19a.html} \end{APACrefURL}
\PrintBackRefs{\CurrentBib}

\bibitem [\protect \citeauthoryear {%
Van Den~Oord%
, Vinyals%
\BCBL {}\ \protect \BOthers {.}}{%
Van Den~Oord%
\ \protect \BOthers {.}}{%
{\protect \APACyear {2017}}%
}]{%
van2017neural}
\APACinsertmetastar {%
van2017neural}%
\begin{APACrefauthors}%
Van Den~Oord, A.%
, Vinyals, O.%
\BCBL {}\ \BOthersPeriod {.}\end{APACrefauthors}%
\unskip\
\newblock
\APACrefYearMonthDay{2017}{}{}.
\newblock
{\BBOQ}\APACrefatitle {Neural discrete representation learning} {Neural discrete representation learning}.{\BBCQ}
\newblock
\APACjournalVolNumPages{Advances in neural information processing systems}{30}{}{}.
\PrintBackRefs{\CurrentBib}

\bibitem [\protect \citeauthoryear {%
Wang%
, Bovik%
, Sheikh%
\BCBL {}\ \BBA {} Simoncelli%
}{%
Wang%
\ \protect \BOthers {.}}{%
{\protect \APACyear {2004}}%
}]{%
wang2004image}
\APACinsertmetastar {%
wang2004image}%
\begin{APACrefauthors}%
Wang, Z.%
, Bovik, A\BPBI C.%
, Sheikh, H\BPBI R.%
\BCBL {}\ \BBA {} Simoncelli, E\BPBI P.%
\end{APACrefauthors}%
\unskip\
\newblock
\APACrefYearMonthDay{2004}{}{}.
\newblock
{\BBOQ}\APACrefatitle {Image quality assessment: from error visibility to structural similarity} {Image quality assessment: from error visibility to structural similarity}.{\BBCQ}
\newblock
\APACjournalVolNumPages{{IEEE} Trans. Image Process.}{13}{4}{600--612}.
\newblock
\begin{APACrefURL} \url{https://doi.org/10.1109/TIP.2003.819861} \end{APACrefURL}
\newblock
\begin{APACrefDOI} \doi{10.1109/TIP.2003.819861} \end{APACrefDOI}
\PrintBackRefs{\CurrentBib}

\bibitem [\protect \citeauthoryear {%
Wang%
\ \BBA {} Ji%
}{%
Wang%
\ \BBA {} Ji%
}{%
{\protect \APACyear {2022}}%
}]{%
wang2022open}
\APACinsertmetastar {%
wang2022open}%
\begin{APACrefauthors}%
Wang, Z.%
\BCBT {}\ \BBA {} Ji, H.%
\end{APACrefauthors}%
\unskip\
\newblock
\APACrefYearMonthDay{2022}{}{}.
\newblock
{\BBOQ}\APACrefatitle {Open Vocabulary Electroencephalography-to-Text Decoding and Zero-Shot Sentiment Classification} {Open vocabulary electroencephalography-to-text decoding and zero-shot sentiment classification}.{\BBCQ}
\newblock
\BIn{} \APACrefbtitle {Thirty-Sixth {AAAI} Conference on Artificial Intelligence, {AAAI} 2022, Thirty-Fourth Conference on Innovative Applications of Artificial Intelligence, {IAAI} 2022, The Twelveth Symposium on Educational Advances in Artificial Intelligence, {EAAI} 2022 Virtual Event, February 22 - March 1, 2022} {Thirty-sixth {AAAI} conference on artificial intelligence, {AAAI} 2022, thirty-fourth conference on innovative applications of artificial intelligence, {IAAI} 2022, the twelveth symposium on educational advances in artificial intelligence, {EAAI} 2022 virtual event, february 22 - march 1, 2022}\ (\BPGS\ 5350--5358).
\newblock
\APACaddressPublisher{}{{AAAI} Press}.
\newblock
\begin{APACrefURL} \url{https://doi.org/10.1609/aaai.v36i5.20472} \end{APACrefURL}
\newblock
\begin{APACrefDOI} \doi{10.1609/AAAI.V36I5.20472} \end{APACrefDOI}
\PrintBackRefs{\CurrentBib}

\bibitem [\protect \citeauthoryear {%
Xi%
\ \protect \BOthers {.}}{%
Xi%
\ \protect \BOthers {.}}{%
{\protect \APACyear {2023}}%
}]{%
xi2023unicorn}
\APACinsertmetastar {%
xi2023unicorn}%
\begin{APACrefauthors}%
Xi, N.%
, Zhao, S.%
, Wang, H.%
, Liu, C.%
, Qin, B.%
\BCBL {}\ \BBA {} Liu, T.%
\end{APACrefauthors}%
\unskip\
\newblock
\APACrefYearMonthDay{2023}{}{}.
\newblock
{\BBOQ}\APACrefatitle {UniCoRN: Unified Cognitive Signal ReconstructioN bridging cognitive signals and human language} {Unicorn: Unified cognitive signal reconstruction bridging cognitive signals and human language}.{\BBCQ}
\newblock
\APACjournalVolNumPages{arXiv preprint arXiv:2307.05355}{}{}{}.
\PrintBackRefs{\CurrentBib}

\bibitem [\protect \citeauthoryear {%
Xia%
, Yin%
\BCBL {}\ \BBA {} Li%
}{%
Xia%
\ \protect \BOthers {.}}{%
{\protect \APACyear {2024}}%
}]{%
xia2024decoding}
\APACinsertmetastar {%
xia2024decoding}%
\begin{APACrefauthors}%
Xia, R.%
, Yin, C.%
\BCBL {}\ \BBA {} Li, P.%
\end{APACrefauthors}%
\unskip\
\newblock
\APACrefYearMonthDay{2024}{}{}.
\newblock
{\BBOQ}\APACrefatitle {Decoding the Echoes of Vision from fMRI: Memory Disentangling for Past Semantic Information} {Decoding the echoes of vision from fmri: Memory disentangling for past semantic information}.{\BBCQ}
\newblock
\BIn{} \APACrefbtitle {Proceedings of the 2024 Conference on Empirical Methods in Natural Language Processing} {Proceedings of the 2024 conference on empirical methods in natural language processing}\ (\BPGS\ 2040--2052).
\PrintBackRefs{\CurrentBib}

\bibitem [\protect \citeauthoryear {%
Xu%
, Wang%
, Zhang%
, Wang%
\BCBL {}\ \BBA {} Shi%
}{%
Xu%
\ \protect \BOthers {.}}{%
{\protect \APACyear {2023}}%
}]{%
xu2023versatile}
\APACinsertmetastar {%
xu2023versatile}%
\begin{APACrefauthors}%
Xu, X.%
, Wang, Z.%
, Zhang, G.%
, Wang, K.%
\BCBL {}\ \BBA {} Shi, H.%
\end{APACrefauthors}%
\unskip\
\newblock
\APACrefYearMonthDay{2023}{}{}.
\newblock
{\BBOQ}\APACrefatitle {Versatile diffusion: Text, images and variations all in one diffusion model} {Versatile diffusion: Text, images and variations all in one diffusion model}.{\BBCQ}
\newblock
\BIn{} \APACrefbtitle {Proceedings of the IEEE/CVF International Conference on Computer Vision} {Proceedings of the ieee/cvf international conference on computer vision}\ (\BPGS\ 7754--7765).
\PrintBackRefs{\CurrentBib}

\bibitem [\protect \citeauthoryear {%
Yin%
, Ye%
\BCBL {}\ \BBA {} Li%
}{%
Yin%
\ \protect \BOthers {.}}{%
{\protect \APACyear {2024}}%
}]{%
yin2024language}
\APACinsertmetastar {%
yin2024language}%
\begin{APACrefauthors}%
Yin, C.%
, Ye, Z.%
\BCBL {}\ \BBA {} Li, P.%
\end{APACrefauthors}%
\unskip\
\newblock
\APACrefYearMonthDay{2024}{}{}.
\newblock
{\BBOQ}\APACrefatitle {Language Reconstruction with Brain Predictive Coding from fMRI Data} {Language reconstruction with brain predictive coding from fmri data}.{\BBCQ}
\newblock
\APACjournalVolNumPages{arXiv preprint arXiv:2405.11597}{}{}{}.
\PrintBackRefs{\CurrentBib}

\bibitem [\protect \citeauthoryear {%
Zhang%
, Han%
, Worth%
\BCBL {}\ \BBA {} Liu%
}{%
Zhang%
\ \protect \BOthers {.}}{%
{\protect \APACyear {2020}}%
}]{%
zhang2020connecting}
\APACinsertmetastar {%
zhang2020connecting}%
\begin{APACrefauthors}%
Zhang, Y.%
, Han, K.%
, Worth, R.%
\BCBL {}\ \BBA {} Liu, Z.%
\end{APACrefauthors}%
\unskip\
\newblock
\APACrefYearMonthDay{2020}{}{}.
\newblock
{\BBOQ}\APACrefatitle {Connecting concepts in the brain by mapping cortical representations of semantic relations} {Connecting concepts in the brain by mapping cortical representations of semantic relations}.{\BBCQ}
\newblock
\APACjournalVolNumPages{Nature communications}{11}{1}{1877}.
\PrintBackRefs{\CurrentBib}

\end{thebibliography}

\end{document}